\documentclass[conference]{IEEEtran}

\usepackage{times}
\usepackage{epsfig}
\usepackage{graphicx,wrapfig}
\usepackage{amsmath}
\usepackage{amssymb}
\usepackage{booktabs}
\usepackage{multirow}
\usepackage[dvipsnames,table,xcdraw]{xcolor}

\usepackage[breaklinks=true,bookmarks=false]{hyperref}
\usepackage[capitalize]{cleveref}

\begin{document}

\title{ClipSitu: Effectively Leveraging CLIP for Conditional Predictions in Situation Recognition}

\author{\IEEEauthorblockN{Debaditya Roy\IEEEauthorrefmark{1},
Dhruv Verma\IEEEauthorrefmark{1}\IEEEauthorrefmark{2}, and
Basura Fernando\IEEEauthorrefmark{1}\IEEEauthorrefmark{2}}
\IEEEauthorrefmark{1} Institute of High Performance Computing (IHPC)\\ Agency for Science, Technology and Research (A*STAR)\\
1 Fusionopolis Way, \#16-16 Connexis, Singapore 138632, Republic of Singapore\\
\IEEEauthorrefmark{2}
Centre for Frontier AI Research (CFAR)\\ Agency for Science, Technology and Research (A*STAR)\\  1 Fusionopolis Way, \#16-16 Connexis, Singapore 138632, Republic of Singapore
}

\maketitle

\begin{abstract}
Situation Recognition is the task of generating a structured summary of what is happening in an image using an activity verb and the semantic roles played by actors and objects.
In this task, the same activity verb can describe a diverse set of situations as well as the same actor or object category can play a diverse set of semantic roles depending on the situation depicted in the image.  
Hence a situation recognition model needs to understand the context of the image and the visual-linguistic meaning of semantic roles. 
Therefore, we leverage the CLIP foundational model that has learned the context of images via language descriptions. 
We show that deeper-and-wider multi-layer perceptron (MLP) blocks obtain noteworthy results for the situation recognition task by using CLIP image and text embedding features and it even outperforms the state-of-the-art CoFormer, a Transformer-based model, thanks to the external implicit visual-linguistic knowledge encapsulated by CLIP and the expressive power of modern MLP block designs. 
Motivated by this, we design a cross-attention-based Transformer using CLIP visual tokens that model the relation between textual roles and visual entities.  
Our cross-attention-based Transformer known as ClipSitu XTF outperforms existing state-of-the-art by a large margin of 14.1\% on semantic role labelling (value) for top-1 accuracy using imSitu dataset. 
{Similarly, our ClipSitu XTF obtains state-of-the-art situation localization performance.}
We will make the code publicly available.
\end{abstract}

\section{Introduction}\label{sec:intro}

Situation recognition was first introduced to computer vision in pioneering work~\cite{yatskar2016situation}. 
Situation recognition is an important problem in scene understanding, activity understanding, and action reasoning as it provides a structured representation of the main activity depicted in the image. 
The key component in situation recognition is the task of semantic role labeling.  
Semantic role labeling is complex as the same activity verb may have different functional meanings and purposes depending on the context of the image.
For example, the verb ``spray" can be used to describe a firefighter spraying water on a fire, someone spraying oil on salad, someone spraying perfume on their face, and someone spraying hairspray on their hair.
Hence, semantic role labeling requires a detailed understanding of the event in the image using contextual information from the image and how it relates to the linguistic definition of the event in terms of the activity (verb) and activity-specific roles.

Multimodal Foundation Models such as CLIP \cite{radford2021learning} and ALIGN \cite{jia2021scaling} provide context as they are trained on many millions of image/text pairs to capture cross-modal dependencies between images and text. 
In these millions of examples, CLIP model might encounter different usages of the same verb that describe visually different but semantically similar images. 
Hence, CLIP is an excellent multimodal foundation model for solving image semantic role labeling tasks as it provides a grounded understanding of visual and linguistic information.
In \cite{doveh2022teaching}, CLIP is shown to be trainable for complex vision and language tasks termed Structured Vision and Language Concepts.
Another way to leverage multimodal foundation models is to apply an MLP on top of the image encoder in works such as VL-Adapter \cite{sung2022vl}, AIM \cite{yangaim}, EVL \cite{lin2022frozen} and wise-ft \cite{wortsman2022robust}.
These approaches can be applied for predicting the main activity in the image i.e. for image classification \cite{wortsman2022robust} or action detection \cite{yangaim}.
However, semantic role labeling is a conditional classification task that needs verb and role along with the image.
Therefore, in~\cite{li2022clip}, authors convert situation recognition to a text-prompt-based prediction problem by fine-tuning a CLIP image encoder with the text outputs from a large language model -- GPT-3 \cite{brown2020language} called CLIP-Event.
The verbs are ranked using the prompt ``An image of ⟨verb⟩'' based on image CLIP embeddings.
After predicting the verb, each noun is predicted using another text prompt “The ⟨name⟩ is a ⟨role⟩ of ⟨verb⟩'', i.e. ``The firefighter is an agent in spraying''.
Even with the world knowledge in GPT-3, CLIP-Event performs worse on semantic role labeling than state-of-the-art CoFormer \cite{cho2022collaborative} which is directly trained on the images.
The reason is that finetuning CLIP on semantic role labeling is not effective as the dataset imSitu \cite{yatskar2016situation} is not massive containing only 126,102 images yet it contains a massive amount of nouns (11,538) that are related to 190 unique roles. Therefore, the mapping between roles and nouns becomes an extremely challenging task.

We show that a well-designed multimodal MLP that consists of a modern MLP block design is able to solve semantic role labeling using CLIP embeddings and it outperforms the state-of-the-art without finetuning the CLIP model.
This multimodal MLP is trained on a combination of image and text embedding from the verb and the role obtained from the CLIP model. 
Multimodal MLP predicts the entity corresponding to the role using a simple loss function.
Motivated by the effectiveness of CLIP-based multimodal MLP, we adopt a Transformer encoder to leverage the connection across semantic roles in an image. 
Each semantic role is represented using a multimodal input of image and text embedding of the verb and the role.
We show that sharing information across semantic roles using a Transformer leads to slightly improved performance.
{Through the multimodal MLP and Transformer we find that CLIP-based image, verb and role embeddings are effective for role prediction and predicting all roles for a verb by sharing information across them further improves the efficacy.}
Motivated by these two findings, we design a cross-attention Transformer to learn the relation between semantic role queries and CLIP-based visual token representations of the image to further enhance the connection between visual and textual entities.
We term this model as ClipSitu XTF and it obtains state-of-the-art results for Situation Recognition on imSitu dataset outperforming state-of-the-art CoFormer \cite{cho2022collaborative} by 14.1\% on top-1 value performance.
{Similarly, we leverage the cross-attention scores to localize the role in the image. 
Using ClipSitu XTF, we obtain state-of-the-art results for situation localization.}

\section{Related Work}

\textbf{Situation Recognition.} To understand the relationship between different entities in an image, tasks such as image captioning \cite{ke2019reflective, kim2021vilt, guo2022images}, scene graph generation \cite{xu2017scene,cong2021spatial}, and human-object interaction detection \cite{gkioxari2018detecting,lim2023ernet} have been proposed in the literature.
In situation recognition~\cite{yatskar2016situation}, the situational verbs and their roles are obtained based on the meaning of the activity in each image from FrameNet \cite{fillmore2003background}. The entities for each role are populated using the large object dataset ImageNet. 
Recently, situation recognition has also been extended to videos with the VidSitu dataset \cite{sadhu2021visual} where each video spans multiple events each of which is described using a situational verb, semantic roles, and their nouns. 
The VidSitu dataset is extended with grounded entities in \cite{khangrounded} while \cite{xiao2022hierarchical} proposes a contrastive learning objective framework for video semantic role labeling.
We limit the scope of this work to situation recognition in images.

\textbf{One-stage prediction} approaches predict the situational verb from the image and then the nouns associated with the roles of those verbs.
In \cite{yatskar2016situation}, a conditional random field model is proposed that decomposes the task of situation recognition into verb prediction and semantic role labeling (SRL). 
For SRL, they optimize the log-likelihood of the ground-truth nouns corresponding to each role for an image over possible semantic role-noun pairs from the entire dataset.
In \cite{yatskar2017commonly}, a tensor decomposition model is used on top of CRF that scores combinations of role-noun pairs. 
They also perform semantic augmentation to provide extra training samples for rarely observed noun-role combinations.
In \cite{mallya2017recurrent}, a predefined order for semantic roles is decided to predict the nouns for an image, and a recurrent neural network is used to predict the nouns in that order. 
Authors in \cite{li2017situation} propose a gated graph neural network (GGNN) to capture all possible relations between roles instead of a predefined order as in \cite{mallya2017recurrent}.
In \cite{suhail2019mixture}, a mixture kernel
is applied to relate the nouns predicted for one role with respect to the noun predicted for another role. 
These relations provide a prior for the GGNN \cite{li2017situation} to predict nouns.

In \cite{pratt2020grounded}, imSitu is extended with grounded entities in each image to create Situations With Grounding (SWiG) dataset. 
They propose two models -- Independent Situation Localizer (ISL) and Joint Situation Localizer (JSL). 
Both ISL and JSL use LSTMs to predict nouns in a predefined sequential order similar to \cite{mallya2017recurrent} while RetinaNet estimates the locations of entities.
A transformer encoder-decoder architecture is proposed in \cite{cho2021gsrtr} where the encoder 
captures semantic features from the image for verb prediction and the decoder learns the role relations.
In \cite{jiangexploiting}, situational verbs are predicted using a CLIP encoder on the image and the  detected objects in the image.

\textbf{Two-stage prediction} approaches introduce  an additional stage to enhance the verb prediction using the predicted nouns of the roles.
In \cite{cooray2020attention}, transformers are used to predict semantic roles using interdependent queries that contain the context of all roles.  
The context acts as the key and values while the verb and the role form the query to predict the noun.
They also consider the nouns of two predefined roles along with the image to enhance the verb prediction using a CNN. 
In \cite{wei2022rethinking}, a coarse-to-fine refinement of verb prediction is proposed by re-ranking verbs based on the nouns predicted for the roles of the verb. 
CoFormer\cite{cho2022collaborative} combines ideas from \cite{wei2022rethinking} and \cite{cho2021gsrtr} with transformer encoder and decoder predicting verbs and nouns, respectively. 
They add another encoder-decoder to refine the verb prediction based on the decoder outputs from the noun decoder.

\section{CLIPSitu Models and Training}

In this section, first, we present how we extract CLIP~\cite{radford2021learning} embedding (features) for situation recognition.
{After that we present the verb prediction model. 
Then, we present three models for Situation Recognition using the CLIP embeddings. }
Afterward, we present a loss function that we use to train our models.

\subsection{Extracting CLIP embedding}
Every image $I$ has a situational action associated with it, denoted by a verb $v$. 
For this verb $v$, there is a set of semantic roles $R_v = \{r_1, r_2, \cdots, r_m\}$ each of which is played by an entity denoted by its noun value $N = \{ n_1, n_2, \cdots, n_m\}$.
We use CLIP~\cite{radford2021learning} visual encoder $\psi_v()$, and the text encoder $\psi_t()$ to obtain representations for the image, verb, roles, and nouns denoted by $X_I$, $X_V$, $X_{R_v}$ and, $X_N$ respectively.
Here $X_{R_v} = \{ X_{r_1}, X_{r_2}, \cdots, X_{r_m}\}$ {for $m$ roles} and $X_N = \{ X_{n_1}, X_{n_2}, \cdots, X_{n_m} \}$ {for corresponding $m$ nouns} where $X_{r_i} = \psi_t(r_i)$ and $X_{n_i} = \psi_t(n_i)$ are obtained using text encoder. Similarly, the $X_I = \psi_v(I)$ and $X_V = \psi_t(v)$ is obtained using vision encoder. Note that all representations $X_I, X_V, X_{r_i}$ and $X_{n_i}$ have the same dimensions.
In imSitu dataset\cite{yatskar2016situation}, we have 504 unique verbs, 190 unique roles, and 11538 unique nouns.
We extract CLIP text embeddings each verb, role, and noun separately.

\subsection{ClipSitu Verb MLP}
The first task in situation recognition is to predict the situational verb correctly from the image.
We design a simple MLP with CLIP embeddings of the image $X_I$ as input called ClipSitu Verb MLP as follows:
\begin{equation}
    \hat{v} = \phi_{V}(X_I).
\end{equation}
where $\phi_V$ contains $l$ linear layers of a fixed dimension with ReLU activation to predict the situational verb.
Just before the final classifier, there is a Dropout layer with a 0.5 rate.
We train ClipSitu Verb MLP with standard cross-entropy loss.

\subsection{ClipSitu MLP}
\label{sec.meth.cmlp}
Here we present a modern multimodal MLP block design for semantic role labeling for Situation Recognition that predicts each semantic role of a verb in an image.
We term this method as \textbf{ClipSitu MLP}.
Specifically, given the image, verb, and role embedding, the ClipSitu MLP predicts the embedding of the corresponding noun value for the role.
In contrast to what has been done in the literature, ClipSitu MLP obtains contextual information by conditioning the information from the image, verb, and role embeddings.
While the image embedding provides context about the possible nouns for the role, the verb provides the context on how to interpret the image situation.
\begin{figure}
    \centering
    \includegraphics[width=\linewidth]{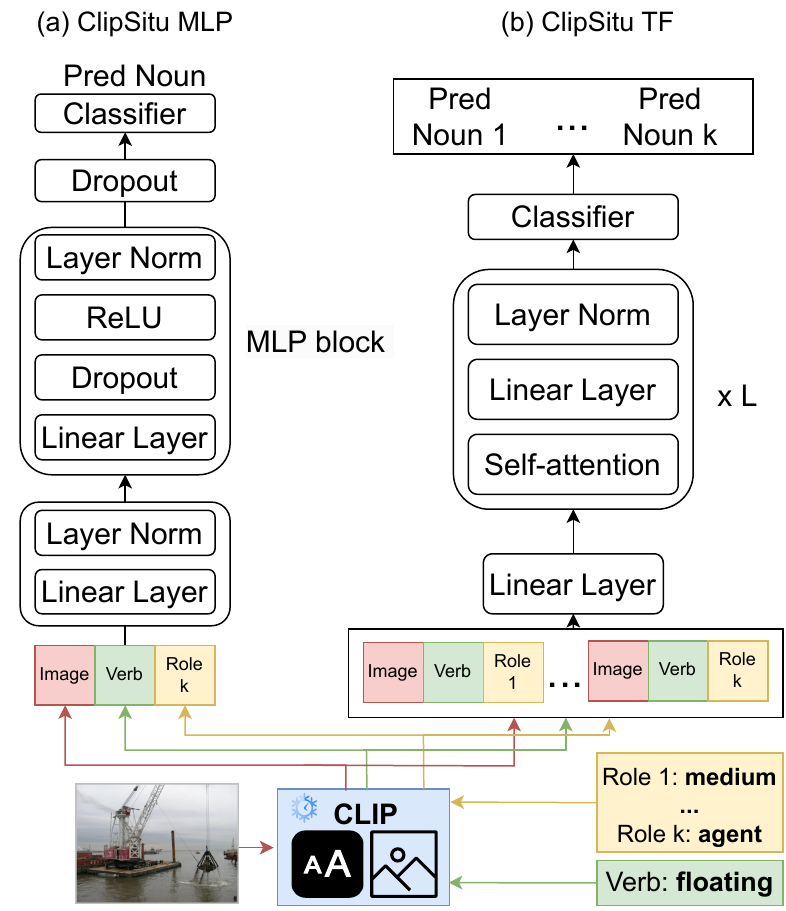}
    \caption{{Architecture of the ClipSitu MLP and TF models. We use pooled image embedding from the CLIP image encoder for ClipSitu MLP and TF. In ClipSitu TF, all the roles for the verb are predicted simultaneously.}}
    \label{fig:mlp_tf}
\end{figure}

We concatenate the role embedding for each role ${r_i}$ to the image and verb embedding to form the multimodal input $X_i$ where $X_i = [X_I, X_v, X_{r_i}]$. 
Then, we stack $l$ MLP blocks to construct CLIPSitu MLP and use it to transform the multimodal input $X_i$ to predict the noun embedding $\hat{X}_{n_i}$ as follows:
\begin{equation}
    \hat{X}_{n_i} = \phi_{MLP}(X_i).
\end{equation}
In $\phi_{MLP}$, the first MLP block projects the input feature $X_i$ to a fixed hidden dimension using a linear projection layer followed by a LayerNorm~\cite{ba2016layer}.
Each subsequent MLP block consists of a Linear layer followed by a Dropout layer (with a dropout rate of 0.2), ReLU~\cite{nair2010rectified}, and a LayerNorm as shown in \cref{fig:mlp_tf}(a).
We predict the noun class from the predicted noun embedding using a dropout layer (rate 0.5) followed by a linear layer which we name as classifier $\phi_c$ as 
\begin{equation}
    \hat{y}_{n_i} =\texttt{argmax}~ \phi_c(\hat{X}_{n_i})
\end{equation}
where $\hat{y}_{n_i}$ is the predicted noun class.
We use cross-entropy loss between predicted $\hat{y}_{n_i}$ and ground truth nouns ${y}_{n_i}$ as explained later in~\cref{loss} to train the model.

\subsection{ClipSitu TF: ClipSitu Transformer}\label{sec:tf}
The role-noun pairs associated with a verb in an image are related as they contribute to different aspects of the execution of the verb. 
Hence, we extend our ClipSitu MLP model using a Transformer~\cite{vaswani2017attention} to exploit the interconnected semantic roles and predict them in parallel.
The input to the Transformer is similar to ClipSitu MLP (i.e. $X_i = [X_I, X_v, X_{r_i}]$), however, we build a set of vectors using $\{ X_1, X_2, \cdots, X_m\}$ where $m$ denotes the number of roles of the verb.
Each vector in the set is further processed by a linear projection to reduce dimensions.
We initialize a Transformer model $\phi_{TF}$ with $l$ encoder layers and multi-head attention with $h$ heads.
Using the Transformer model, we predict the value embedding of the $m$ roles as output tokens of the transformer
\begin{equation}
   \{ \hat{X}_{n_1}, \hat{X}_{n_2}, \cdots, \hat{X}_{n_m}\} = \phi_{TF}(\{ X_1, X_2, \cdots, X_m\}).
\end{equation}
Similar to the MLP, we predict the noun classes using a classifier on the value embedding as $\hat{y}_i = \texttt{argmax}~ \phi_c(\hat{X}_{n_i})$ where $ i= \{1, \cdots, m\}$ as shown in \cref{fig:mlp_tf}(b).

\begin{figure}
    \centering
    \includegraphics[width=0.9\linewidth]{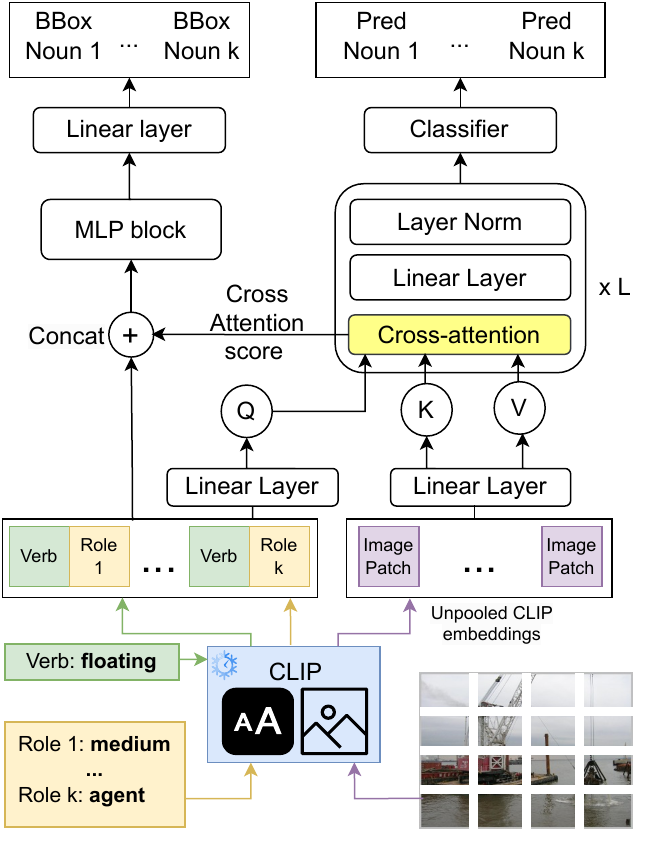}
    \caption{{Architecture of ClipSitu XTF. We use embeddings from each patch of the image obtained from CLIP image encoder.}}
    \label{fig:xtf}
\end{figure}

\subsection{ClipSitu XTF: Cross-Attention Transformer}
Each semantic role in a situation is played by an object located in a specific region of the image.
Therefore, it is important to pay attention to
the regions of the image which has a stronger relationship with the role.
Such a mechanism would allow us to obtain better noun prediction accuracy.
Hence, we propose to use the encoding for each patch of the image obtained from the CLIP model.
We design a cross-attention Transformer called \textbf{ClipSitu XTF} to model how each patch of the image is related to every role of the verb through attention as shown in \cref{fig:xtf}.

Let the patch embedding of an image be denoted by $X_{I,p} = \{ X^1_I, X^2_I, \cdots, X^p_I\}$ where $p$ is the number of image patches.
These patch embeddings form the key and values of the cross-attention Transformer while the verb-role embedding is the query in Transformer. 
The verb embedding is concatenated with each role embedding to form $m$ verb-role embeddings $X_{vr} = \{ [X_V; X_{r_1}], [X_V; X_{r_2}], \cdots, [X_V; X_{r_m}] \}$.
We project each verb-role embedding to the same dimension as the image patch embedding using a linear projection layer.
Then the cross-attention operator in a Transformer block is denoted as follows:
\begin{align}
Q = W_Q &X_{vr}, K = V = W_I X_{I,p} \nonumber\\
&\hat{X} = softmax\frac{QK^T}{\sqrt{d_K}} V
\end{align}
where $W_Q$ and $W_I$ represent projection weights for queries, keys, and values and $d_K$ is the dimension of the key token $K$. 
As with ClipSitu TF, we have $l$ cross-attention layers in ClipSitu XTF.   
The predicted output from the final cross-attention layer contains $m$ noun embeddings $\hat{X} =  \{ \hat{X}_{n_1}, \hat{X}_{n_2}, \cdots, \hat{X}_{n_m}\}$. 
Similar to the transformer in \cref{sec:tf}, we predict the noun classes using a classifier on the noun embeddings as $\hat{y}_i = \phi_c(\hat{X}_{n_i})$ where $ i= \{1, \cdots, m\}$.

{Next, we use ClipSitu XTF to perform localization of roles that requires predicting a bounding box $\textbf{b}_i$ for every role $r_i$ in the image.
The cross-attention scores from the first layer of ClipSitu XTF $A_{m \times p}$ are rearranged into $m$ score vectors $\{A_1, \cdots, A_m\}$. 
Each score vector $A_i$ is $p$-dimensional and shows how each patch in the image is related to the verb and role. 
To incorporate the verb and role context to the score, we concatenate each score vector to its corresponding verb-role embedding from $X_{vr}$ to obtain input for localization $X_l = \{[X_V; X_{r_1}; A_1], [X_V; X_{r_2}; X_2], \cdots, [X_V; X_{r_m}; X_m] \}$.
We pass $X_l$ through a single MLP block (designed for ClipSitu MLP) followed by a linear layer and a sigmoid function to obtain the predicted bounding box $\hat{b}_i \in [0, 1]^4$ for every role $r_i$. 
The four elements in the predicted bounding box indicate the center coordinates, height and width relative to the input image size. 
Though ClipSitu XTF uses cross-attention as CoFormer \cite{cho2022collaborative}, the verb role tokens in the query and the image tokens are obtained from CLIP and not learned.
Leveraging the power of CLIP embeddings allows us to design a simpler one-stage ClipSitu XTF compared to the two-stage CoFormer \cite{cho2022collaborative}.
}

\subsection{Losses}
\label{loss}
\textbf{Minimum Annotator Cross Entropy Loss.} The imSitu dataset employs three annotators to label each noun for a role. 
In some instances, annotators may not provide the same annotation.
Existing approaches \cite{cooray2020attention, cho2022collaborative} make multiple predictions instead of one to tackle this issue.
However, this can confuse the network during training as for there are multiple annotations for the same example.
Furthermore, the loss function should not penalize a prediction that is close to any of the annotators' ground truth but further away from others. 
We propose minimum cross-entropy loss that considers the ground truth from each annotator. 
For a prediction $\hat{y}_i$, ground truth from all the annotators $\mathcal{A}= \{ A_1, \cdots A_q \}$ is used to train our network as follows
\begin{equation}
    \mathcal{L}_{MAXE} = \min_{\mathcal{A}} -\sum_{c=1}^C y^{(A_j)}_{i,c} log(\hat{y}_{i,c}) \text{ where } \forall A_j \in \mathcal{A}. 
\end{equation}
Here, $C$ denotes the total number of classes and $\mathcal{L}_{MAXE}$ stands for Minimum Annotator Cross Entropy Loss. {To train ClipSitu XTF for localization of roles, we employ $L1$ loss to compare the predicted and ground-truth bounding boxes 
\begin{equation}
    \mathcal{L}_{L1} = \frac{1}{m}\sum^m_{i=1} \| b_i - \hat{b}_i \|_1.
\end{equation}
We train ClipSitu XTF for noun prediction and localization using the combined loss $\mathcal{L} = \mathcal{L}_{MAXE} + \mathcal{L}_{L1}$. 
}

\section{Experiments}

\subsection{Evaluation Details}
{We perform our experiments on imSitu dataset \cite{yatskar2016situation} and the augmented imSitu dataset called SWiG \cite{pratt2020grounded} for situation recognition and localization, respectively termed as grounded situation recognition.}
The dataset has a total of 125k images with 75k train, 25k validation, and 25k test images.
The metrics used for semantic role labeling are \textit{value} and \textit{value-all} \cite{yatskar2016situation} which predict the accuracy of noun prediction for a given role.
For a given verb with $k$ roles, \textit{value} measures whether the predicted noun for at least one of $k$ roles is correct.
On the other hand, \textit{value-all} measures whether all the predicted nouns for all $k$ roles are correct.
A prediction is correct if it matches the annotation of any one of the three annotators. 
{Situation localization metrics \textit{grnd value} and \textit{grnd value-all} compute the accuracy of bounding box prediction \cite{pratt2020grounded} similar to value and value-all.
A predicted bounding box is correct if the overlap with the ground truth is $\ge$ 0.5.
}
{The metrics value, value-all, grnd value and grnd value-all are evaluated in three settings based on whether we are using ground truth verb, top-1 predicted verb, or top-5 predicted verbs.
For our model ablation on semantic role labeling and situation localization, we use the ground truth verb setting for measuring value, value-all, grnd value, and grnd value-all.
All experiments are performed on the dev set unless otherwise specified.}

\subsection{Implementation Details}
We use the CLIP model with ViT-B32 image encoder to extract image features unless otherwise specified.  
The input to ClipSitu MLP is a concatenation of the CLIP embeddings of the image, verb, and role, each of 512 dimensions leading to 1536 dimensions. 
For both ClipSitu TF and XTF, we set the sequence length to be 6 which refers to the maximum number of roles possible for a verb following \cite{cooray2020attention}.
Each verb has a varying number of roles and we mask the inputs that are not required.
For ClipSitu TF, each input token in the sequence is the concatenated image, verb, and role CLIP embedding same as the MLP above which is projected to 512 dimensions using a linear layer.
For the patch-based cross-attention Transformer (ClipSitu XTF), we obtain the embedding for input image patches from CLIP image encoder (ViT-B32 model) which results in 50 tokens ($224/32 \times 224/32$ + 1 class) of 512 dimensions that are used as key and value. 
The query tokens are concatenated verb and role CLIP embeddings that are projected to 512 dimensions using a linear layer.
Unless otherwise mentioned, we train all our models with a batch size of 64, a learning rate of 0.001, and  an ExponentialLR scheduler with Adamax optimizer, on a 24 GB Nvidia 3090.

\subsection{Verb and noun prediction with different CLIP Image Encoders}
In \cref{tab:verbablation}, we compare the proposed ClipSitu Verb MLP model against {zero-shot and linear probe performance of CLIP}. 
We also compare against a state-of-the-art CLIP finetuning model called weight-space ensembles (wise-ft) \cite{wortsman2022robust} that leverages both zero-shot and fine-tuned CLIP models to make verb predictions.
We compare ClipSitu Verb MLP and wise-ft using 4 CLIP image encoders - ViT-B32, ViT-B16, ViT-L14, and ViT-L14@336px.
The image clip embeddings for ViT-B32 and ViT-B16 are 512 dimensions and for ViT-L14, and ViT-L14@336px are 768 dimensions.
These four encoders represent different image patch sizes, different depths of image transformers, and different input image sizes.
The hidden layer is 1024 dimensional in the ClipSitu Verb MLP.
{Zero-shot performance of CLIP suggests that CLIP image features are beneficial for situation recognition tasks.}  
Increasing the number of hidden layers does not improve performance for ClipSitu Verb MLP as it obtains the best top-1 and top-5 verb prediction even with a single hidden layer.
{ClipSitu Verb MLP performs better than wise-ft and linear probe for all image encoders which shows that MLP based finetuning on CLIP image features works better than finetuning the CLIP image encoder itself or using regression (linear probe) for situational verb prediction.
Our best performing ClipSitu Verb MLP outperforms linear probe by 4.46\% on Top-1 and wise-ft by 3.2\% on Top-5 when using the same ViT-L14 image encoder.}

\begin{table}
\centering
\scriptsize
\begin{tabular}{l|l|c|c|c} 
\hline
\begin{tabular}[c]{@{}l@{}}Image\\ Encoder\end{tabular} & Verb Model & \begin{tabular}[c]{@{}c@{}}Hidden\\ Layer\end{tabular} & Top-1 & Top-5 \\ 
\hline
\multirow{6}{*}{ViT-B32} & {zero-shot} & {-} & {29.20} & {65.21} \\ 
\cline{2-5}
 & {linear probe} & {-} & {44.63} & {78.35} \\ 
\cline{2-5}
 & wise-ft & - & 46.51 & 74.30 \\ 
\cline{2-5}
 & \multirow{3}{*}{ClipSitu Verb MLP} & 1 & 46.69 & 76.11 \\
 &  & 2 & 46.51 & 76.08 \\
 &  & 3 & 44.51 & 74.15 \\ 
\hline
\multirow{6}{*}{ViT-B16} & {zero-shot} & {-} & {31.90} & {67.89} \\ 
\cline{2-5}
 & {linear probe} & {-} & {49.27} & {78.76} \\ 
\cline{2-5}
 & wise-ft & - & 48.77 & 83.45 \\ 
\cline{2-5}
 & \multirow{3}{*}{ClipSitu Verb MLP} & 1 & 50.91 & 89.57 \\
 &  & 2 & 50.83 & 89.40 \\
 &  & 3 & 48.63 & 88.55 \\ 
\hline
\multirow{4}{*}{ViT-L14} & {zero-shot} & {-} & {38.18} & {79.34} \\ 
\cline{2-5}
 & {linear probe} & {-} & {52.39} & {87.67} \\ 
\cline{2-5}
 & wise-ft & - & 51.51 & 84.30 \\ 
\cline{2-5}
 & \multirow{3}{*}{ClipSitu Verb MLP} & 1 & 56.70 & 84.61 \\
 &  & 2 & 56.63 & 84.49 \\
 &  & 3 & 53.80 & 82.44 \\ 
\hline
\multirow{6}{*}{\begin{tabular}[c]{@{}l@{}}ViT-L14\\@336px\end{tabular}} & {zero-shot} & {-} & {39.70} & {79.21} \\ 
\cline{2-5}
 & {linear probe} & {-} & {53.40} & {81.45} \\ 
\cline{2-5}
 & wise-ft & - & 52.22 & 82.95 \\ 
\cline{2-5}
 & \multirow{3}{*}{ClipSitu Verb MLP} & 1 & \textbf{57.86} & \textbf{86.11} \\
 &  & 2 & 56.22 & 84.55 \\
 &  & 3 & 54.35 & 82.85 \\
\hline
\end{tabular}
\caption{Comparing performance of ClipSitu Verb MLP with zero-shot and finetuned CLIP (linear probe and wise-ft \cite{wortsman2022robust}).}
\label{tab:verbablation}
\end{table}

\begin{table}[t]
\centering
\scriptsize
\begin{tabular}{l|l|cc|cc|cc}
\hline
\multirow{2}{*}{\begin{tabular}[c]{@{}l@{}}Image\\ Encoder\end{tabular}} & \multirow{2}{*}{\begin{tabular}[c]{@{}c@{}} Model\end{tabular}} & \multicolumn{2}{c|}{Top-1 } & \multicolumn{2}{c|}{Top-5} & \multicolumn{2}{c}{Ground truth} \\ \cline{3-8} 
 & & value & v-all & value & v-all & value & v-all \\ \hline
{\multirow{3}{*}{ViT-B32}} & {MLP} & 45.65 & 27.06 &  66.27 & 37.55 & 76.91 & 43.22 \\
{} & {TF} & 45.67 & 27.33 & 66.28 & 37.98 & 76.77 & 42.97 \\
{} & {XTF} & 44.54 & 25.94 & 64.93 & 35.56 & 75.25 & 40.79 \\ \hline
{\multirow{3}{*}{ViT-B16}} & {MLP} & 46.33 & 28.29 & 67.37 & 39.45 & 77.88 & 44.78 \\
{} & {TF} & 46.41 & 28.65 & 67.39 & 39.75 & 77.23 & 43.82 \\
{} & {XTF} & 45.67 & 27.44 & 66.09 & 37.42 & 75.43 & 40.58 \\ \hline
{\multirow{3}{*}{ViT-L14}} & { MLP} & 46.46 & 28.39 & 67.61 & 39.71 & 77.63 & 43.94 \\
{} & {TF} & 46.95 & 29.56 & 68.19 & 41.22 & 78.02 & 45.25 \\
{} & {XTF} &  46.95 & 29.49 & 68.08 & 40.61 & 77.84 & 44.54 \\ \hline
\multirow{3}{*}{\begin{tabular}{@{}l@{}}ViT-L14\\@336px\end{tabular}} & {MLP} & 46.74  & 29.06 & 67.90 & 40.54 & 77.93 & 44.88 \\
{} & {TF} & 46.97  & 29.66 & 68.27 & 41.41 & 78.30 & 45.79 \\
{} &  XTF & \textbf{47.17} & \textbf{30.06} & \textbf{68.44} & \textbf{41.66} & \textbf{78.49} & \textbf{45.81} \\ \hline
\end{tabular}
    \caption{Comparison of CLIP Image Encoders on noun prediction task using top-1 and top-5 predicted verb from  the best-performing Verb MLP model obtain from ~\cref{tab:verbablation}. All models' performance improves by increasing the number of patch tokens either by reducing patch size (32$\rightarrow$16$\rightarrow$14) or increasing image size (224$\rightarrow$336). v-all stands for value all.
    } 
    \label{tab:clipablation}
\end{table}

\begin{table*}
\centering
\scriptsize
\resizebox{\linewidth}{!}{
\begin{tabular}{ll|ccccc|ccccc|cccc}
\hline
 &  & \multicolumn{5}{c|}{Top-1 predicted verb} & \multicolumn{5}{c|}{Top-5 predicted verb} & \multicolumn{4}{c}{Ground truth verb} \\ \cline{3-16} 
\multirow{-2}{*}{Set} & \multirow{-2}{*}{Method} & verb & value & value-all & { {\begin{tabular}[c]{@{}l@{}}grnd\\ value\end{tabular}}}& { {\begin{tabular}[c]{@{}l@{}}grnd\\ value-all\end{tabular}}} & verb & value & value-all & { {\begin{tabular}[c]{@{}l@{}}grnd\\ value\end{tabular}}}  & { {\begin{tabular}[c]{@{}l@{}}grnd\\ value-all\end{tabular}}} & value & value-all & { {\begin{tabular}[c]{@{}l@{}}grnd\\ value\end{tabular}}} & { {\begin{tabular}[c]{@{}l@{}}grnd\\ value-all\end{tabular}}} \\ \hline
 & CRF \cite{yatskar2016situation} & {32.25} & {24.56} & 14.28 & { -} & { -} & {58.64} & {42.68} & 22.75 & { -} & { -} & {65.90} & 29.50 & { -} & { -} \\
 & CRF w/ DataAug \cite{yatskar2017commonly} & {34.20} & {26.56} & 15.61 & { -} & { -} & {62.21} & {46.72} & 25.66 & { -} & { -} & {70.80} & 34.82 & { -} & { -} \\
 & RNN w/ Fusion \cite{mallya2017recurrent} & {36.11} & {27.74} & 16.60 & { -} & { -} & {63.11} & {47.09} & 26.48 & { -} & { -} & {70.48} & 35.56 & { -} & { -} \\
 & GraphNet \cite{li2017situation} & {36.93} & {27.52} & 19.15 & { -} & { -} & {61.80} & {45.23} & 29.98 & { -} & { -} & {68.89} & 41.07 & { -} & { -} \\
 & CAQ w/ RE-VGG \cite{cooray2020attention} & {37.96} & {30.15} & 18.58 & { -} & { -} & {64.99} & {50.30} & 29.17 & { -} & { -} & {73.62} & 38.71 & { -} & { -} \\
 & Kernel GraphNet \cite{suhail2019mixture} & {43.21} & {35.18} & 19.46 & { -} & { -} & {68.55} & {56.32} & 30.56 & { -} & { -} & {73.14} & 41.68 & { -} & { -} \\
 & ISL \cite{pratt2020grounded} & {38.83} & {30.47} & 18.23 & { 22.47} & { 07.64} & {65.74} & {50.29} & 28.59 & { 36.90} & { 11.66} & {72.77} & 37.49 & { 52.92} & { 15.00} \\
 & JSL \cite{pratt2020grounded} & {39.60} & {31.18} & 18.85 & { 25.03} & { 10.16} & {67.71} & {52.06} & 29.73 & { 41.25} & { 15.07} & {73.53} & 38.32 & { 57.50} & { 19.29} \\
 & GSRTR \cite{cho2021gsrtr} & {41.06} & {32.52} & 19.63 & { 26.04} & { 10.44} & {69.46} & {53.69} & 30.66 & { 42.61} & { 15.98} & {74.27} & 39.24 & { 58.33} & { 20.19} \\
 & SituFormer \cite{wei2022rethinking} & 44.32 & 35.35 & 22.10 & { 29.17} & { 13.33} & 71.01 & 55.85 & 33.38 & { 45.78} & { 19.77} & 76.08 & 42.15 & { \textbf{61.82}} & { 24.65} \\
 & CoFormer \cite{cho2022collaborative} & {44.41} & {35.87} & 22.47 & { 29.37} & { 12.94} & {72.98} & {57.58} & 34.09 & { 46.70} & { 19.06} & {76.17} & 42.11 & { 61.15} & { 23.09} \\ \cline{2-16}
\multirow{-12}{*}{dev} & ClipSitu XTF & \multirow{1}{*}{\textbf{58.19}} & \textbf{47.23} & \textbf{29.73} & {\textbf{41.30} } & { \textbf{13.92}} & \multirow{1}{*}{\textbf{85.69}} & \textbf{68.42} & \textbf{41.42} & {\textbf{49.23}} & {\textbf{23.45}} & \textbf{78.52} & \textbf{45.31} &  
{55.36} & {\textbf{32.37}} \\ \hline

 & CRF \cite{yatskar2016situation} & {32.34} & {24.64} & 14.19 & { -} & { -} & {58.88} & {42.76} & 22.55 & { -} & { -} & {65.66} & 28.96 & { -} & { -} \\
 & CRF w/ DataAug \cite{yatskar2017commonly} & {34.12} & {26.45} & 15.51 & { -} & { -} & {62.59} & {46.88} & 25.46 & { -} & { -} & {70.44} & 34.38 & { -} & { -} \\
 & RNN w/ Fusion \cite{mallya2017recurrent} & {35.90} & {27.45} & 16.36 & { -} & { -} & {63.08} & {46.88} & 26.06 & { -} & { -} & {70.27} & 35.25 & { -} & { -} \\
 & GraphNet \cite{li2017situation} & {36.72} & {27.52} & 19.25 & { -} & { -} & {61.90} & {45.39} & 29.96 & { -} & { -} & {69.16} & 41.36 & { -} & { -} \\
 & CAQ w/ RE-VGG \cite{cooray2020attention} & {38.19} & {30.23} & 18.47 & { -} & { -} & {65.05} & {50.21} & 28.93 & { -} & { -} & {73.41} & 38.52 & { -} & { -} \\
 & Kernel GraphNet \cite{suhail2019mixture} & {43.27} & {35.41} & 19.38 & { -} & { -} & {68.72} & {55.62} & 30.29 & { -} & { -} & {72.92} & 42.35 & { -} & { -} \\
 & ISL \cite{pratt2020grounded} & {39.36} & {30.09} & 18.62 & { {22.73}} & { {07.72}} & {65.51} & {50.16} & 28.47 & { {36.6}} & { {11.56}} & {72.42} & 37.10 & { {52.19}} & { {14.58}} \\
 & JSL \cite{pratt2020grounded} & {39.94} & {31.44} & 18.87 & { {24.86}} & { {09.66}} & {67.60} & {51.88} & 29.39 & { {40.6}} & { {14.72}} & {73.21} & 37.82 & { {56.57}} & { {18.45}} \\
 & GSRTR \cite{cho2021gsrtr} & {40.63} & {32.15} & 19.28 & { {25.49}} & { {10.10}} & {69.81} & {54.13} & 31.01 & { {42.5}} & { {15.88}} & {74.11} & 39.00 & { {57.45}} & { {19.67}} \\
 & SituFormer \cite{wei2022rethinking} & 44.20 & 35.24 & 21.86 & { {29.22}} & { {13.41}} & 71.21 & 55.75 & 33.27 & { {46.00}} & { {20.10}} & 75.85 & 42.13 & { \textbf{61.89}} & { {24.89}} \\
 & CoFormer \cite{cho2022collaborative} & {44.66} & {35.98} & 22.22 & { {29.05}} & { {12.21}} & {73.31} & {57.76} & 33.98 & { {46.25}} & { {18.37}} & {75.95} & 41.87 & { {60.11}} & { {22.12}} \\
 & CLIP-Event \cite{li2022clip} & 45.60 & 33.10 & 20.10 & { 21.60} & {10.60} & - & - & - & {-} & { -} & - & - & { -} & { -} \\ \cline{2-16} 
 \multirow{-13}{*}{test} & ClipSitu XTF & \multirow{1}{*}{\textbf{58.19}} & \textbf{47.23} & \textbf{29.73} & {\textbf{40.01}} & {\textbf{15.03}} & 
 \multirow{1}{*}{\textbf{85.69}} & \textbf{68.42} & 
 \textbf{41.42} & {\textbf{49.78}} & {\textbf{25.22}} & \textbf{78.52} & \textbf{45.31} & {54.36} & {\textbf{33.20}} \\ \hline
\end{tabular}
}
\caption{Comparison with state-of-the-art on {Grounded Situation Recognition}. Robustness of ClipSitu MLP, TF, and XTF is demonstrated by the massive improvement for value and value-all with Top-1 and Top-5 predicted verbs over SOTA. 
}
\label{tab:sotatable}

\end{table*}

Next, we study the effect of using different CLIP image encoders for noun prediction with ClipSitu MLP, TF and XTF.
We compare ViT-B32, ViT-B16, ViT-L14, and ViT-L14@336px.
For ClipSitu XTF, the number of image patch tokens used as key and value changes based on patch size and image size.
We have 197 tokens ($224/16 \times 224/16$ + 1 class token) for ViT-B16, 257 tokens for ViT-L14 ($224/14 \times 224/14$ + 1 class token), and 
577 tokens for ViT-L14@336px ($336/14 \times 336/14$ + 1 class token).
For ViT-L14, and ViT-L14@336px image encoders, we obtain 768-dimensional embeddings which are projected using a linear layer to 512.
We choose the best hyperparameters for ClipSitu MLP, TF, and XTF whose ablations are presented in \cref{sec:archabl}.

In \cref{tab:clipablation}, we observe that the value and value-all using ground truth verbs steadily improve for all three models as the number of patches increases from 32 to 16 to 14 or the image size increases from 224 to 336.
For ViT-B32 and ViT-B16, the best performance is obtained by ClipSitu MLP but it drops with ViT-L14.
On the other hand, the maximum improvement is seen in ClipSitu XTF i.e. 5.1\% for value-all compared to 1.6\% and 2.8\% for ClipSitu MLP and TF, respectively. 
ClipSitu XTF is able to extract more relevant information when attending to more image patch tokens to produce better predictions.
To compare noun prediction using top-1 and top-5 predicted verbs, we use the best ClipSitu Verb MLP (ViT-L14@336px) from \cref{tab:verbablation}.
For both Top-1 and Top-5 predicted verbs, we observe a similar trend as the ground truth verb. 
ClipSitu XTF again shows the most improvement in value and value-all to obtain the best performance among the three models across ground truth, Top-1 and Top-5 predicted verbs.

\subsection{Comparison with SOTA}

In \cref{tab:sotatable}, we compare the performance of proposed approaches with state-of-the-art approaches on situation recognition.
We use ViT-L14@336px image encoder for all models -- ClipSitu Verb MLP, ClipSitu MLP, ClipSitu TF, and ClipSitu XTF.
ClipSitu Verb MLP outperforms SOTA method CoFormer on Top-1 and Top-5 verb prediction by a large margin of 12.6\% and 12.4\%, respectively, on the test set, which shows the effectiveness of using CLIP image embeddings over directly predicting the verb from the images.
The comparison with existing works shows that with a  well-designed MLP network, ClipSitu MLP outperforms state-of-the-art CoFormer \cite{cho2022collaborative} in all metrics comprehensively.
ClipSitu MLP, TF, and XTF also handily outperform the only other CLIP-based semantic role labeling method, CLIP-Event \cite{li2022clip}.
ClipSitu XTF performs the best for noun prediction based on both the predicted top-1 verb and top-5 verb for value and value-all matrices.
ClipSitu XTF outperforms state-of-the-art CoFormer by a massive margin of 14.1\% on top-1 value and by 9.6\% on top-1 value-all using the Top-1 predicted verb on the test set. 
{Furthermore, on situation localization, ClipSitu XTF performs significantly better than state-of-the-art for top-1 grnd value by 11\% while showing improvements on all metrics.}
%

\subsection{Ablations on hyperparameters}\label{sec:archabl}
\begin{table*}[t]
\resizebox{\linewidth}{!}{
\begin{tabular}{ll|cccc|cccc|cccc|cccc}
\hline
Heads &  & \multicolumn{4}{c|}{1} & \multicolumn{4}{c|}{2} & \multicolumn{4}{c|}{4} & \multicolumn{4}{c}{8}\\ \hline
Layers &  & 1 & 2 & 4 & 6 & 1 & 2 & 4 & 6 & 1 & 2 & 4 & 6 & 1 & 2 & 4 & 6\\ \hline
\multirow{2}{*}{ClipSitu TF} & value & 75.73 & 75.78 & \textbf{76.87} & 24.68 & 75.80 & 75.95 & 75.97 & 18.28 & 75.71 & 76.77 & 75.87 & 05.20 & 75.74 & 75.93 & 76.07 & 75.94\\
 & value-all & 41.40 & 41.52 & 41.84 & 00.21 & 41.64 & 41.60 & 41.83 & 00.21 & 41.43 & 42.10 & 41.75 & 00.00 & 41.35 & 41.58 & \textbf{42.97} & 41.72\\ 
 \hline
\multirow{2}{*}{ClipSitu XTF} & value & 72.70 & 74.33 &  \textbf{75.27} & 74.35 & 53.11 & 53.17 & 53.11 & 53.16 & 53.18 & 53.51 & 53.45 & 53.49 & 53.13 & 53.44 & 53.38 & 53.54\\
 & value-all & 36.61 & 39.11 & \textbf{40.79} & 39.06 & 16.58 & 16.64 & 16.38 & 16.46 & 16.50 & 16.77 & 16.90 & 16.97 & 16.42 & 16.89 & 16.85 & 17.02 \\
\hline
\end{tabular}}
\caption{Ablation on Transformer hyperparameters. 1 head with 4 layers is sufficient to obtain best value and value-all performance for ClipSitu XTF. For TF, 1 head and 4 layers produces best value whereas 8 heads and 4 layers produces best value-all performance.}
\label{tab:tfablation}
\end{table*}
\begin{figure}
    \begin{tabular}{l}
    \includegraphics[width=0.48\linewidth]{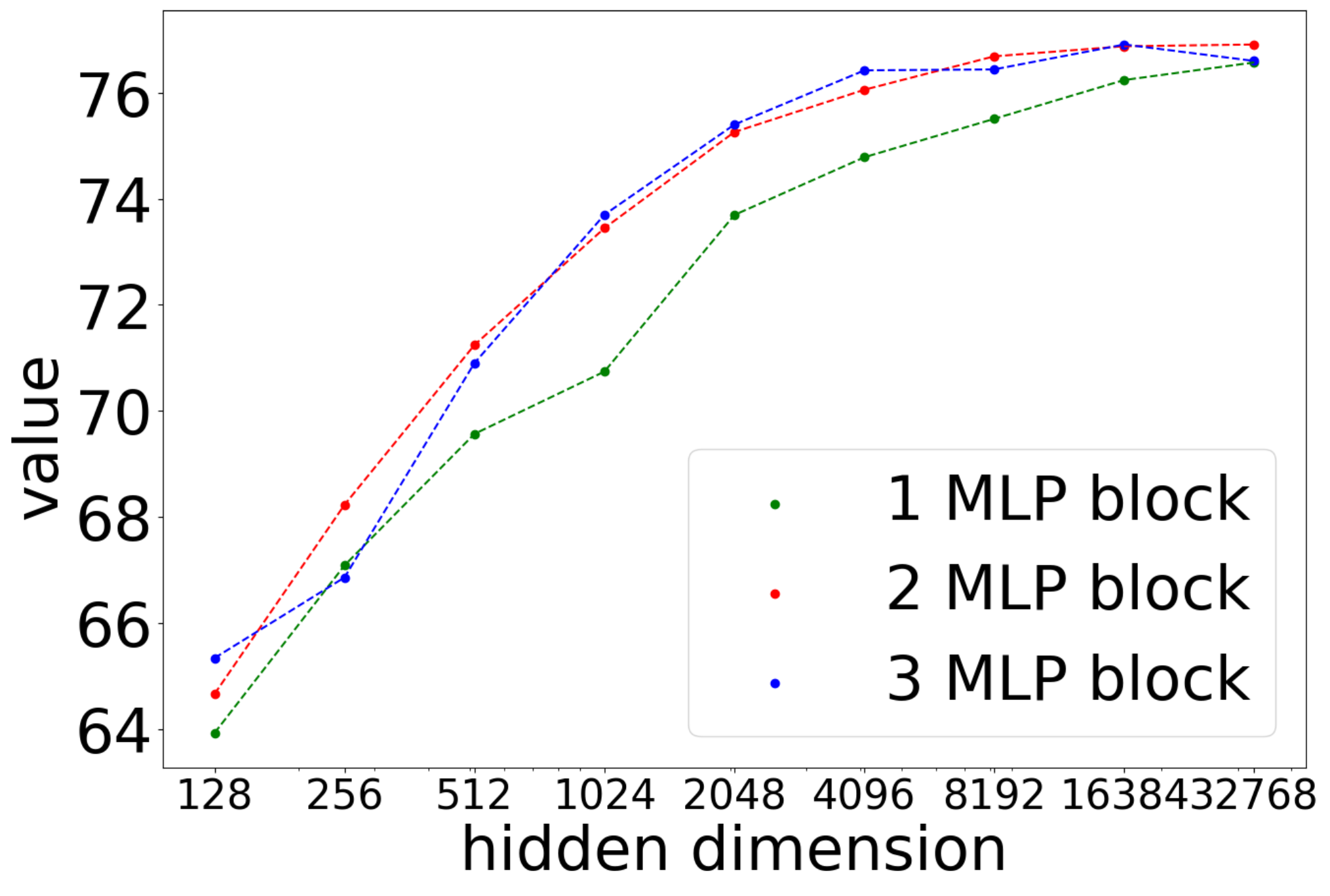}  \includegraphics[width=0.48\linewidth]{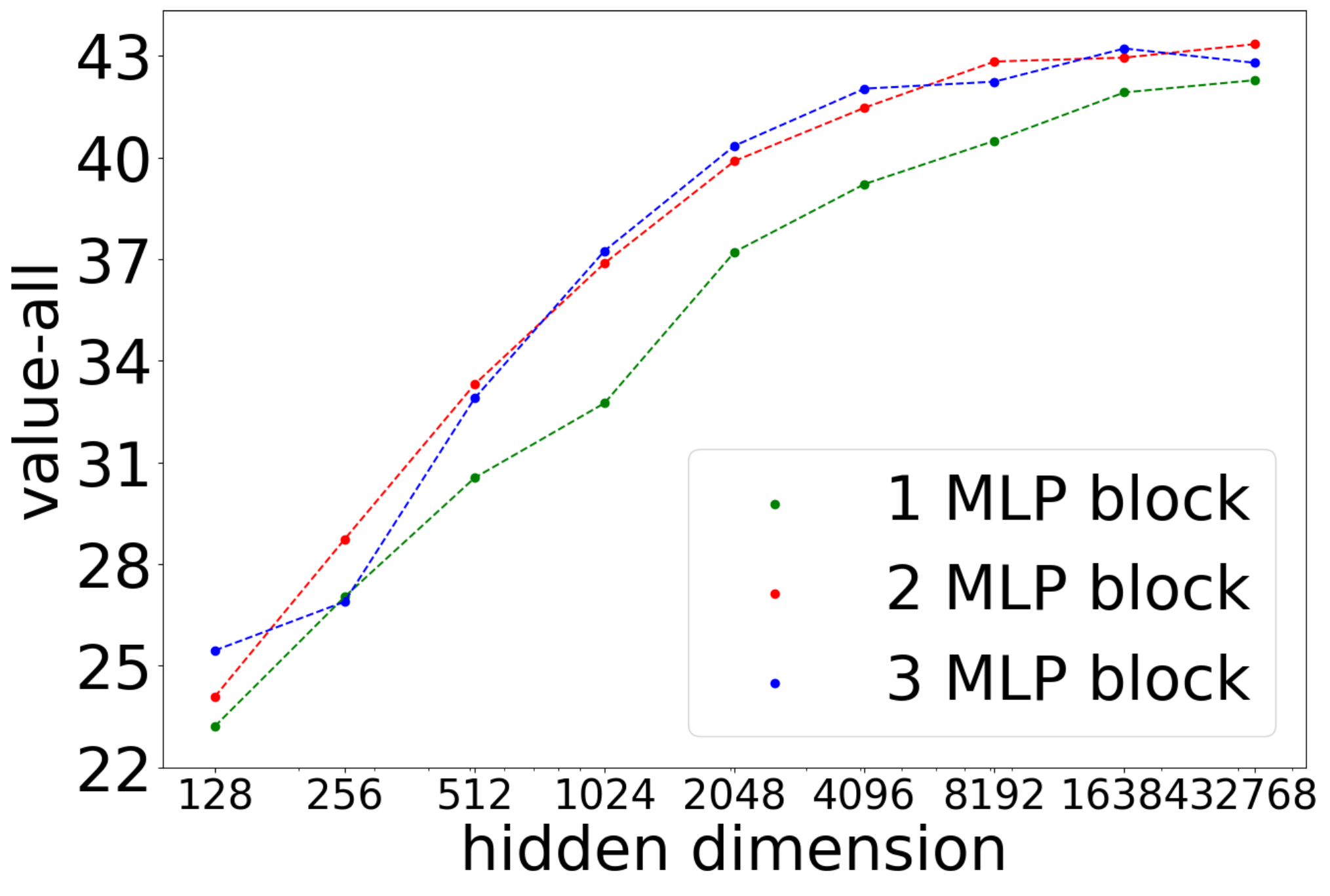} \\
    \end{tabular}
    \caption{Effect of the number of MLP blocks and hidden dimensions on value and value-all. We train with very large hidden dimensions such as 8192, 16384, and 32768 to obtain state-of-the-art value and value-all results. }
    \label{fig:mlpablation}
\end{figure}
\begin{table}
\scriptsize
\centering
\begin{tabular}{l|cc|cc|cc}
\hline
\multirow{2}{*}{\begin{tabular}[c]{@{}c@{}} {Model}\end{tabular}} & \multicolumn{2}{c|}{{Top-1} } & \multicolumn{2}{c|}{{Top-5}} & \multicolumn{2}{c}{{Ground truth}} \\ \cline{2-7} 
  & \begin{tabular}[c]{@{}l@{}}{grnd}\\ {value}\end{tabular} & \begin{tabular}[c]{@{}l@{}}{grnd}\\ {v-all}\end{tabular} & \begin{tabular}[c]{@{}l@{}}{grnd}\\ {value}\end{tabular}& \begin{tabular}[c]{@{}l@{}}{grnd}\\ {v-all}\end{tabular} & \begin{tabular}[c]{@{}l@{}}{grnd}\\ {value}\end{tabular} & \begin{tabular}[c]{@{}l@{}}{grnd}\\ {v-all}\end{tabular} \\ \hline
  {XAtt. L1} & {39.30} & {10.54} & {46.46} & {19.70} & {53.87} & {30.22} \\
  {XAtt. L4} & {{33.30} } & { {09.34}} & {{42.55}} & {{17.53}} & 
{{51.32}} & {{31.32}}\\
  \begin{tabular}[c]{@{}l@{}}{XAtt. L1  + L4} \end{tabular} & {{36.56} } & { {09.88}} & {{44.23}} & {{11.43}} & 
{{54.56}} & {{34.71}}\\
  \begin{tabular}[c]{@{}l@{}}{XAtt L1 +} \\{verb-role emb}\end{tabular} & {\textbf{41.30} } & { \textbf{13.92}} & {\textbf{49.23}} & {\textbf{23.45}} & 
{\textbf{55.36}} & {\textbf{32.37}}\\ \hline
 \end{tabular}
        \caption{{Comparing different ClipSitu XTF inputs for situation localization. XAtt. -- cross-attention scores. L1 -- first XTF layer and L4 -- last XTF layer of best performing model (1 head, 4 layers, ViT-L14@336px). Concatenating (+) verb role embedding (emb) improves the performance of cross-attention scores. We use the best Verb MLP from \cref{tab:verbablation}. v-all stands for value-all.}}
    \label{tab:grndcompare}
\end{table}

In \cref{fig:mlpablation}, we explore combinations of MLP blocks and the hidden dimensions of each block to obtain the best MLP network for semantic role labeling. 
Increasing the number of MLP blocks and hidden dimensions steadily improves performance as the number of unique nouns to be predicted is 11538.
We train MLP with small to very large hidden dimensions i.e. 128$\rightarrow$16384 which results in a steady improvement in both value and value-all.
No improvement in value and value-all is seen when we increase the layer dimension further to 32768 for 3 MLP blocks which demonstrates that we have reached saturation.
Our best ClipSitu MLP for semantic role labeling obtains 76.91 for value and 43.22 for value-all with 3 MLP blocks with each block having 16,384 hidden dimensions which beats the state-of-the-art CoFormer \cite{cho2022collaborative}.
The main reason our ClipSitu MLP performs so well on semantic role labeling is our modern MLP block design that contains large hidden dimensions along with LayerNorm which have not been explored in existing MLP-based CLIP finetuning approaches.
We also compare the performance of ClipSitu MLP with the proposed minimum annotator cross-entropy loss ($\mathcal{L}_{MAXE}$) versus applying cross-entropy using the noun labels of each annotator separately.
We find that $\mathcal{L}_{MAXE}$ produces better value and value-all performance (76.91 and 43.22) compared to cross-entropy (76.57 and 42.88).


In \cref{tab:tfablation}, we explore the number of heads and layers needed to obtain the best-performing hyperparameters for semantic role labeling using ClipSitu TF and XTF.
We find that a single head with 4 transformer layers performs the best in terms of value for both ClipSitu TF and XTF while for value-all, an 8-head 4-layer ClipSitu TF performs the best and we use this for subsequent evaluation.
For both ClipSitu TF and XTF, increasing the number of layers beyond 4 does not yield any improvement in value or value-all when using less number of heads (1,2,4).
Similarly, for ClipSitu XTF, increasing the number of heads and layers leads to progressively deteriorating performance.
Both of these performance drops can be attributed to the fact that we have insufficient samples for training larger Transformer networks \cite{radford2021learning}.
{In ClipSitu XTF, we have fixed role tokens obtained from CLIP. 
We found this produces better noun prediction performance than learning the role tokens for each verb. 
Details are in Appendix A.
In \cref{tab:grndcompare}, ablation on situation localization shows that cross-attention scores from the first XTF layer performs better than the last XTF layer. 
We concatenated verb and role embeddings to the cross-attention score of first XTF layer to provide more context about the role which further improves localization performance.}

{\textbf{Complexity.} We compare the number of parameters, computation, and inference time for ClipSitu MLP, TF, and XTF using the ViT-L14-336 image encoder and CoFormer \cite{cho2022collaborative} in \cref{tab:params}.
We find that ClipSitu TF is the most efficient in terms of parameters, computation, and inference time closely followed by ClipSitu XTF at half the parameters of CoFormer and 9\% of ClipSitu MLP.
Even adding the lightweight ClipSitu Verb MLP to ClipSitu XTF for combined verb and noun prediction leads to a very efficient but effective model.
Therefore, we conclude that ClipSitu XTF not only performs the best at semantic role labeling but is also efficient in terms of parameters, computation, and inference time compared to ClipSitu MLP and CoFormer. }
\begin{table}[]
    \centering
    \scriptsize
    \begin{tabular}{l|c|c|c}
    \hline
    Model & \# Parameters & {GFlops} & {Inference Time(ms)}\\
    \hline
         {CoFormer \cite{cho2022collaborative}} & {93.0M} & {1496.67} & 
         {30.62} \\
         ClipSitu Verb MLP & {1.3M}	& {0.17}	& {0.08} \\ 
         ClipSitu MLP &  580.2M & {443.18} & {32.33} \\
         ClipSitu TF & \textbf{20.2M} & {\textbf{8.65}}  &  {\textbf{1.55}}\\
         ClipSitu XTF & 45.3M & {116.01} & {11.17}\\
         \hline
    \end{tabular}
    \caption{Comparison of parameters, flop count and inference time for CoFormer \cite{cho2022collaborative}, Verb MLP, ClipSitu MLP, TF and XTF models.}
    \label{tab:params}
\end{table}

\begin{figure}
    \centering
    \includegraphics[width=\linewidth]{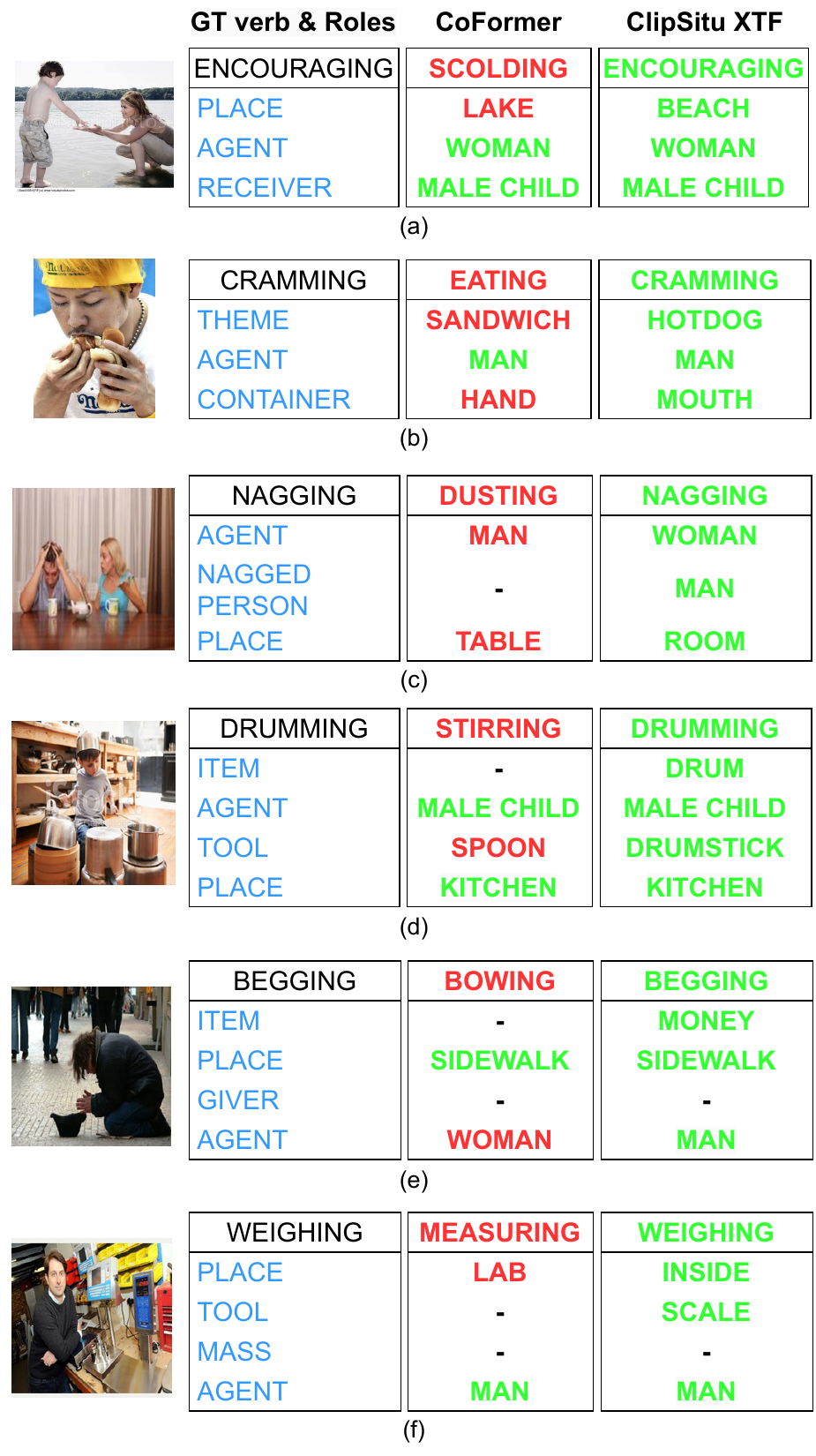}
    \caption{ClipSitu XTF vs. CoFormer \cite{cho2022collaborative} predictions. \textcolor{green}{green} refers to correct prediction while \textcolor{red}{red} refers to incorrect prediction. '-' refers to predicting blank (a noun class) for this role.}
    \label{fig:qual_noun}
\end{figure}

\paragraph{Qualitative Results}
In \cref{fig:qual_noun}, we compare the qualitative results of ClipSitu XTF with CoFormer. 
ClipSitu XTF is able to correctly predicts verbs such as cramming (\cref{fig:qual_noun}(b)) while CoFormer focuses on the action of eating and hence incorrectly predicts the verb which also makes its noun predictions for the container and theme incorrect.
CoFormer predicts the place as table and predicts the verb as dusting (\cref{fig:qual_noun}(c)) instead of focusing on the action of nagging. 
Finally, we see in \cref{fig:qual_noun}(d) that CoFormer is confused by the visual context of kitchen as it predicts stirring instead of identifying the action which is drumming. 
On the other hand, ClipSitu XTF correctly predicts drumming and the tool as drumsticks while still predicting the place as the kitchen. 
{More qualitative results are in Appendix B.}

\section{Conclusion}
We propose to leverage CLIP embeddings for semantic role labeling.
We show that multimodal ClipSitu MLP with large hidden dimensions outperforms the state-of-the-art semantic role labeling approach. 
We propose a ClipSitu XTF model that employs cross-attention between image patch embeddings from the CLIP image encoder and text embeddings.
ClipSitu XTF sets the new state-of-the-art in semantic role labeling improving the current results by a large margin of 14.1\% on top-1 value and by 9.6\% on top-1 value-all.
We also show that our approach of using CLIP embeddings is much more effective than finetuning CLIP, given the relatively small size of the dataset.
Unlike, VL-Adapter \cite{sung2022vl}, AIM \cite{yangaim}, EVL \cite{lin2022frozen} and wise-ft \cite{wortsman2022robust}, our models can handle conditional inputs to solve Situation Recognition task.
Despite the simplicity, our work shows that a traditional approach of freeze and finetune can be still relevant when used with modern neural network designs especially when using Foundational models. 




\bibliographystyle{plain}
\bibliography{clipsitu}

\newpage

\appendix
\section{Learned vs. Fixed role tokens for noun prediction}
We show in \cref{tab:learnvsfixed} that ClipSitu XTF performs worse with learned role tokens for each verb when compared to using fixed role tokens obtained from CLIP.

\begin{table}[]
    \centering
    \resizebox{\linewidth}{!}{
\begin{tabular}{l|cc|cc|cc}
\hline
 \multirow{2}{*}{\begin{tabular}[c]{@{}c@{}} ClipSitu \\ XTF\end{tabular}} & \multicolumn{2}{c|}{Top-1 } & \multicolumn{2}{c|}{Top-5} & \multicolumn{2}{c}{Ground truth} \\ \cline{2-7} 
 & value & v-all & value & v-all & value & v-all \\ \hline
   \begin{tabular}[c]{@{}c@{}} Learnable \\ Role Tokens \end{tabular}  & 44.82	& 25.44	& 65.04	& 35.37	& 75.62	& 44.31 \\ \hline
   \begin{tabular}[c]{@{}c@{}} Fixed \\ Role Tokens \end{tabular} & \textbf{47.17} & \textbf{30.06} & \textbf{68.44} & \textbf{41.66} & \textbf{78.49} & \textbf{45.81} \\ \hline
\end{tabular}}
    \caption{Comparing the performance of noun prediction with fixed and learnable role tokens per verb in ClipSitu XTF.}
    \label{tab:learnvsfixed}
\end{table}

\section{More Qualitative Results}

\begin{figure}[!ht]
    \centering
    \includegraphics[width=0.7\linewidth]{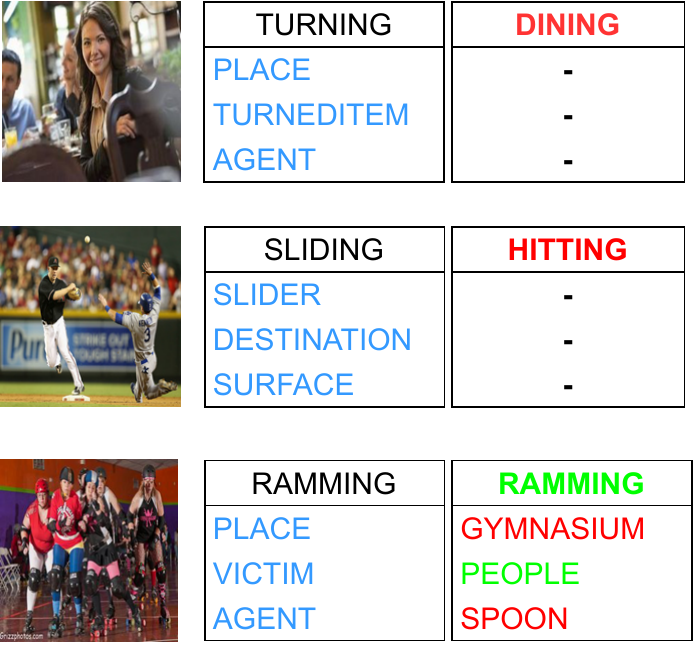}
    \caption{Examples where ClipSitu predicts the verb/noun incorrectly}
    \label{fig:fail}
\end{figure}

In \cref{fig:fail}, we show some qualitative examples where ClipSitu Verb MLP predicts the verb incorrectly or ClipSitu XTF predicts the noun incorrectly. 

\end{document}